\documentclass{article}

\PassOptionsToPackage{numbers, sort&compress}{natbib}  

\usepackage[preprint]{Styles/neurips_2025}

\usepackage[utf8]{inputenc} 
\usepackage[T1]{fontenc}    
\usepackage{hyperref}       
\usepackage{url}            
\usepackage{booktabs}       
\usepackage{amsfonts}       
\usepackage{ulem}           
\usepackage{nicefrac}       
\usepackage{microtype}      
\usepackage{xcolor}         
\usepackage{graphicx}
\usepackage{caption}
\usepackage{lipsum}
\usepackage{makecell}
\usepackage{amsmath}
\usepackage{pifont}
\usepackage{multirow}
\usepackage{float}
\usepackage{array}
\usepackage{subfig}
\usepackage{enumitem}

\definecolor{ForestGreen}{rgb}{0.13, 0.55, 0.13}
\setlist[itemize]{leftmargin=1.3em}
\setlist[enumerate]{leftmargin=1.3em}

\newcolumntype{L}[1]{>{\raggedright\arraybackslash}m{#1}}
\newcolumntype{C}[1]{>{\centering\arraybackslash}m{#1}}

\title{Stabilizing Information Flow: Entropy Regularization for Safe and Interpretable Autonomous Driving Perception}

\author{%
  Haobo Yang, Shiyan Zhang, Zhuoyi Yang, Jilong Guo, Jun Li, Xinyu Zhang\thanks{Corresponding author} \\
  The State Key Laboratory of Automotive Safety and Energy, \\
  School of Vehicle and Mobility, Tsinghua University \\
  \texttt{xyzhang@tsinghua.edu.cn} \\
}

\begin{document}

\maketitle

\begin{abstract}
Deep perception networks in autonomous driving traditionally rely on data-intensive training regimes and post-hoc anomaly detection, often disregarding fundamental information-theoretic constraints governing stable information processing. We reconceptualize deep neural encoders as hierarchical communication chains that incrementally compress raw sensory inputs into task-relevant latent features. Within this framework, we establish two theoretically justified design principles for robust perception: (\textbf{D1}) smooth variation of mutual information between consecutive layers, and (\textbf{D2}) monotonic decay of latent entropy with network depth. Our analysis shows that, under realistic architectural assumptions—particularly blocks comprising repeated layers of similar capacity—enforcing smooth information flow (D1) naturally encourages entropy decay (D2), thus ensuring stable compression. Guided by these insights, we propose \textbf{E\textsubscript{loss}}, a novel entropy-based regularizer designed as a lightweight, plug-and-play training objective. Rather than marginal accuracy improvements, this approach represents a conceptual shift: it unifies information-theoretic stability with standard perception tasks, enabling explicit, principled detection of anomalous sensor inputs through entropy deviations. Experimental validation on large-scale 3D object detection benchmarks (KITTI and nuScenes) demonstrates that incorporating E\textsubscript{loss} consistently achieves competitive or improved accuracy while dramatically enhancing sensitivity to anomalies—amplifying distribution-shift signals by up to two orders of magnitude. This stable information-compression perspective not only improves interpretability but also establishes a solid theoretical foundation for safer, more robust autonomous driving perception systems.
\end{abstract}

\section{Introduction}
\label{sec:intro}

\begin{figure}[t]
  \centering
  \includegraphics[width=\textwidth]{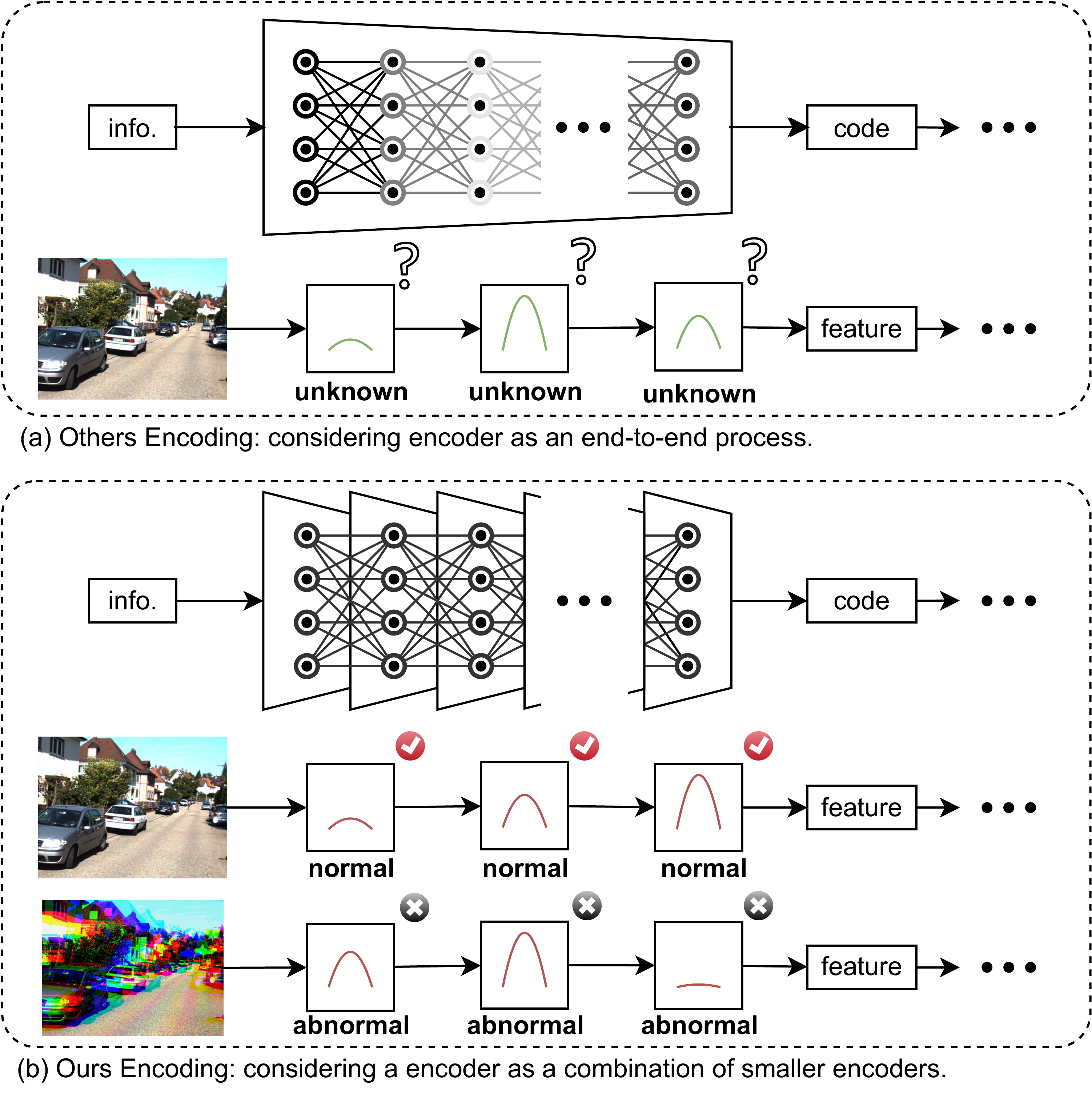}
  \caption{\textbf{Layer-wise view vs.\ end-to-end view.}
  (a)~Conventional pipelines treat the encoder as a single compression block, hiding unstable jumps in information entropy.  
  (b)~Our stable information-compression perspective decomposes the encoder into repeated layers inside each block and \textbf{encourages} smooth entropy decay.  
  Anomalous inputs (dashed red) violate this stability and become easy to spot.}
  \label{fig:layer_stability}
\end{figure}

Intelligent driving promises safer and more efficient urban mobility~\cite{freudendal2019smart}.  
A core enabler is 3D object detection~\cite{zhi2017lightnet,wu2021multi}, whose models must operate under tight latency constraints and withstand harsh, ever-changing road conditions.  
Unlike lab-curated benchmarks, real sensors routinely deliver \textbf{anomalous} frames—e.g.\ under fog, rain, or sensor glitches—that can induce catastrophic perception errors.

Most perception stacks handle anomalies indirectly: they enlarge training sets, inject synthetic noise, or bolt on post-hoc detectors.  
Such data-heavy remedies (i) assume that future outliers resemble past ones and (ii) provide little insight into \textbf{why} the network fails.  
Consequently, accuracy gains on "clean" test sets often fail to translate into robust on-road performance.

\noindent\textbf{Communication view of deep perception.}
We instead model a deep encoder as a \textbf{hierarchical communication chain}~\cite{zou2022novel}: each layer compresses its input into a latent code that is forwarded to the next layer.  
Drawing on source-coding theory~\cite{jones_2000_information}, we articulate two design principles for any well-behaved encoder:

\begin{enumerate}[leftmargin=1.3em]
  \item \label{d1} \textbf{Smooth compression.} The mutual information between successive layers should change gradually.  
  \item \label{d2} \textbf{Monotonic entropy decay.} The entropy of the latent codes should decrease with depth.
\end{enumerate}

Under the common prior that each block is built from repeated layers of similar capacity, enforcing \ref{d1} alone induces an approximately layer-invariant compression ratio, while \ref{d2} is observed to emerge automatically.  
Anomalous frames disrupt this smooth profile (Fig.~\ref{fig:layer_stability}), yielding a principled detection signal.

\noindent\textbf{Contributions.}
We translate these insights into a practical training recipe for robust 3D perception:
\begin{itemize}[leftmargin=1.3em]
  \item \textbf{Stable information-compression framework.}  
        We formalize Principle~\ref{d1} and show analytically that repeated-layer similarity makes entropy decay (Principle~\ref{d2}) arise in expectation.
  \item \textbf{\(E_{\text{loss}}\): a plug-and-play surrogate.}  
        Introducing a continuous latent variable \(X\), we derive \(E_{\text{loss}}\) as the variance penalty on layer-wise entropy drops, fully differentiable and task-agnostic.
  \item \textbf{Accuracy and robustness in practice.}  
        On large-scale autonomous-driving benchmarks, \(E_{\text{loss}}\) (i) matches or exceeds baseline mAP and (ii) boosts anomaly-to-nominal signals by more than two orders of magnitude, with smoother entropy trajectories as an interpretable by-product.
\end{itemize}

This work integrates information-theoretic stability directly into the training objective of 3D perception models and demonstrates its benefits for safety-critical autonomous driving.

\section{Related Works}

\subsection{Uncertainty Quantification}
Autonomous-driving perception must contend with occlusions, sensor noise, and other sources of partial observability that undermine the fidelity of raw measurements \citep{yang2025review}.  
Consequently, quantifying predictive uncertainty has become a pivotal research theme.

\noindent\textbf{Taxonomy of deep-learning uncertainty.}
Following \citep{kendall2017uncertainties}, we distinguish two complementary sources.  
Aleatoric uncertainty captures irreducible noise in the world (for example, weather-induced LiDAR sparsity).  
Because more data cannot eliminate it, practitioners often resort to heteroscedastic modelling; recent work shows that loss terms letting the network predict a variance value can improve robustness \citep{griffiths2021heteroscedastic}.  
Epistemic uncertainty, by contrast, reflects ignorance due to limited data or model capacity and can be reduced given sufficient coverage of the input space.  
Prominent estimators include Monte-Carlo (MC) dropout, which interprets dropout masks as variational samples \citep{goel2021mcdropout}, and deep ensembles, which approximate a posterior with a collection of independent networks \citep{postels2021deterministic}.

\noindent\textbf{Open issues.}
Despite substantial progress, two gaps remain.  
First, most studies lack ground-truth uncertainty labels and therefore evaluate only by correlation with prediction error \citep{barandas2024evaluation}.  
Second, there is no unified quantitative metric usable across classification, regression, and segmentation; task-specific proxies proliferate instead \citep{fukaya2024evaluation}.  
Our work addresses both limitations.  
By casting the encoder as a communication chain with stable information flow, we obtain (i) a binary ground truth (nominal inputs yield stability, anomalies break it) and (ii) a single scalar, \(E_{\text{loss}}\), whose magnitude is comparable across network architectures and learning tasks whenever a repetitive block structure is present.

\subsection{Shannon's Source–Coding View of Neural Encoders}
Classical communication systems enjoy first-principles guarantees from information theory: channel capacity, rate--distortion trade-offs, and provably optimal source codes are all quantified via Shannon entropy \citep{jones_2000_information}.  
Mapping deep networks onto this framework provides a principled lens for both interpretation and optimization.

\noindent\textbf{Neural channels and joint source--channel coding.}
Early work by \citet{mackay_2003_information} cast neuron activations as noisy channels, inspiring a wave of deep joint source--channel coding (JSCC) methods that replace hand-engineered codecs with end-to-end CNNs \citep{kurka_2020_deepjsccf,jankowski_2020_deep}.  
These models demonstrate that learned encoders can match -- or surpass -- Shannon-limit efficiency when trained with appropriate distortion objectives.

\noindent\textbf{Source coding inside representation learning.}
Sharma et al.\ introduced fiducial coding into variational autoencoders, enforcing Shannon's first theorem so that the expected code length tracks the latent entropy \citep{sharma_2021_dagsurv}.  
This alignment guarantees lossless representation under the chosen fidelity criterion.

\noindent\textbf{Information Bottleneck (IB) connection.}
The Information Bottleneck principle \citep{tishby_2000_the} views learning as a compression problem: hidden layers ought to retain only task-relevant information.  
Empirically, multilayer perceptrons first capture mutual information with the target before compressing it away.  
Our work extends this idea by treating each pre-fusion feature extractor as a source coder.  
We explicitly measure the entropy of every layer's output and penalise the variance of the inter-layer entropy change, thereby (i) restricting the optimization search space, (ii) accelerating convergence, and (iii) enabling anomaly detection through violations of the expected entropy profile.

Taken together, these perspectives motivate our stable information-compression framework: by encouraging near-ideal source-coding behaviour inside deep encoders, we inherit the interpretability of communication theory while preserving the expressive power of modern neural networks.

\section{Methodology}
\label{sec:method}

\begin{figure*}[ht!]
  \centering
  \includegraphics[width=0.99\textwidth]{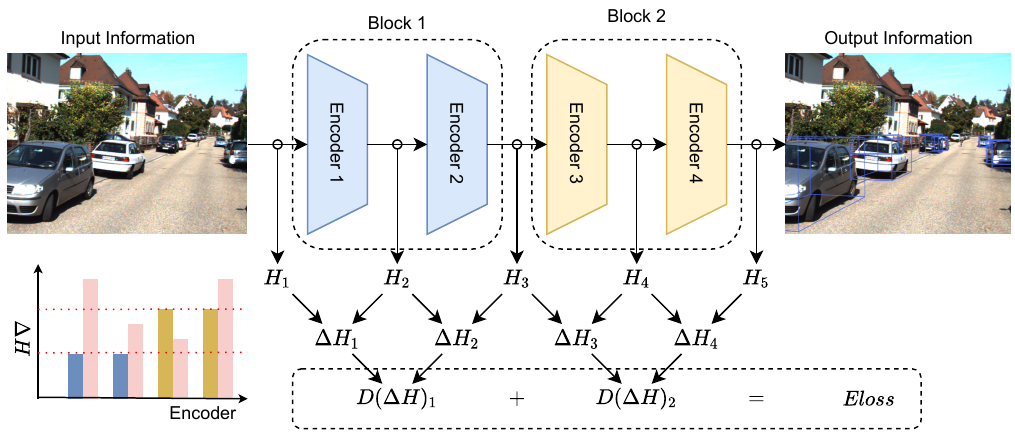}
  \caption{Overview of \textbf{E\textsubscript{loss}}.  
    An input sample passes through two identical encoding blocks, each divided into several near-identical sub-encoders (layers).  
    Within each block we record intermediate representations \(H_n\), compute the entropy drops \(\Delta H_n\), and aggregate them with the variance penalty \(L(\Delta H)\) to obtain a block-level divergence \(D\).  
    Summing the divergences across all blocks yields the global objective
    \(E_{\text{loss}}=\sum_{\text{blocks}} D\).  
    Left: smooth compression inside the blue and yellow blocks; red bars mark unstable layers when \(E_{\text{loss}}\) is absent.}
  \label{fig1}
\end{figure*}

Modern multi-modal perception stacks rely on information compression to discard signals that do not assist the downstream task.  
From a communication-theoretic standpoint, this step is a distortion-constrained source-coding problem: rare source symbols can be pruned to boost transmission efficiency while preserving high reconstruction fidelity at the receiver \citep{jones_2000_information}. This perspective is illustrated in Figure~\ref{fig1}, which outlines the structure and function of our proposed \textbf{E\textsubscript{loss}} mechanism.

\noindent\textbf{Entropy as a proxy for information.}
Shannon entropy quantifies the expected information content of a random variable; lower entropy indicates a more informative code.  
We therefore measure the entropy of each layer’s latent representation and treat its layer-to-layer change as a direct indicator of compression progress.

\noindent\textbf{Stable compression constraint.}
To prevent abrupt distortions, we penalise the variance of the entropy drop across successive layers.  
This encourages a nearly constant compression ratio and improves the efficiency of subsequent fusion and detection modules.

\noindent\textbf{Block granularity and the role of \(D\).}
Our target backbones (for example, SECOND for LiDAR or ResNet-style CNNs) consist of repeated blocks whose internal layer structure is virtually identical.  
For each block \(b\) we compute
\[
D_b = \lambda \, L_b,
\]
where \(L_b\) is the variance penalty derived from that block’s entropy drops.  
The overall regularizer is the sum of these block-level divergences:
\(
E_{\text{loss}} = \sum_b D_b.
\)
Operating at block granularity keeps the stability assumption local and lets \(E_{\text{loss}}\) scale gracefully with network depth.

\noindent\textbf{Comparison with Information Bottleneck.}
Unlike conventional Information Bottleneck (IB) approaches that focus on \textbf{whether} latent features discard irrelevant information, our method emphasizes \textbf{how} the compression proceeds—specifically, its stability and continuity across layers. Traditional IB quantifies the amount of compression between input and output and often relies on variational bounds or adversarial training to approximate mutual information. In contrast, $E_{\text{loss}}$ directly regularizes the variance of layer‐wise entropy drops, resulting in a fully differentiable, structure‐aware objective that plugs into any CNN or Transformer backbone. Moreover, while IB is typically applied to shallow or single‐layer encoders, our approach scales to deep perception networks with repeated sub‐structures, enforcing a globally stable information flow throughout the entire model. This shift—from isolated compression metrics to end‐to‐end compression stability—makes $E_{\text{loss}}$ a lightweight yet principled constraint for modern, safety‐critical perception systems.

The remainder of this section (i) formalizes the entropy estimator, (ii) derives the expression for \(L_b\), and (iii) explains its plug-and-play integration with standard training objectives.

\subsection{Entropy Expectation Across Network Layers}
Neural training can be viewed as a search for a mapping between inputs and targets \citep{wang2024yolov9}.  
Before training, this mapping is weak; during optimization, each layer’s feature extractor—its local compression module—progressively improves.  
Yet the black-box nature of deep models obscures how, or how much, information is actually being compressed \citep{buhrmester2021analysis}.

\noindent\textbf{Source-coding perspective.}
Feature compression corresponds to distortion-constrained source coding: the network should drop only signal components irrelevant to the downstream task.  
Shannon entropy is our proxy for information content.  
In a fixed-bandwidth channel, a steady entropy decrease implies rising transmission efficiency \citep{zou2022novel}.

\noindent\textbf{Layer-wise expectation.}
Many perception backbones comprise repetitive blocks—e.g.\ the stacked linear layers in SECOND \citep{yan2018second}.  
Because each block has near-identical capacity, we treat the internal bandwidth as constant and posit the following expectation:

\begin{quote}
The entropy of successive layer outputs should decay smoothly and monotonically.
\end{quote}

\noindent\textbf{Implication for training.}
By formulating an entropy-based loss that penalises large deviations from this expected decay, we guide optimization toward stable, layer-wise compression.  
This turns an opaque training loop into a process grounded in information theory, yielding both interpretability and improved convergence.

\subsection{Uncertainty Quantification via Entropy Deviations}
When raw sensory data traverse a feature-extraction backbone, each layer produces a new feature map and—ideally—reduces information entropy by filtering out task-irrelevant signals.  
Because no external information is injected, the entropy trajectory should be monotonically decreasing.

\noindent\textbf{Smooth compression under nominal inputs.}
For a block-repetitive network, the entropy drop
\(
\Delta H_n = H_{n+1}-H_n
\)
is roughly proportional to that block’s parameter budget.  
Successive blocks of identical width and structure should therefore exhibit similar \(\Delta H_n\), yielding the stable profile introduced above.

\noindent\textbf{Entropy spikes reveal anomalies.}
Abnormal inputs—e.g.\ corrupted point clouds or sensor glitches—break this regularity.  
Noise injects spurious variance that (i) disrupts the latent code ordering and (ii) can even increase entropy in deeper layers.  
We flag an input as anomalous whenever its entropy sequence
\(\{H_0,H_1,\dots,H_L\}\)
deviates beyond a tolerance band around the expected smooth decay.  
Practically, \(E_{\text{loss}}\) magnifies these deviations: large values correspond to blocks where \(|\Delta H_n|\) diverges from the nominal slope, yielding a unified, architecture-agnostic uncertainty signal.

\subsection{Probabilistic Modelling of Layer Outputs}
Estimating the entropy \(H_n\) of each layer output requires a tractable density model.  
We therefore cast every latent tensor as samples drawn from a continuous random vector.

\noindent\textbf{Feature maps as random vectors.}
Following \citet{zou2022novel}, the \(i\) channels generated by a convolutional kernel form  
\(
\tilde{X}=\{x_1,\dots,x_i\},
\)
independent realisations of a \(d\)-dimensional random variable \(X\), where \(d\) is the number of spatial positions per channel.  
Given \(\tilde{X}\), we fit a simple diagonal-Gaussian density \(p_n(\cdot)\) and compute
\(
\hat{H}_n \;=\; -\mathbb{E}_{x\sim p_n}\bigl[\log p_n(x)\bigr],
\)
an unbiased estimator of the true entropy.

\noindent\textbf{Architecture agnosticism.}
This construction is not limited to CNNs.  
Transformer tokens, point-cloud pillars, or MLP embeddings can be reshaped so that the last dimension indexes channels and the rest form the sample axis; the same estimator then applies, enabling a unified entropy-based loss across backbone families.

\subsection{Entropy Estimation with $k$-Nearest Neighbours}
Although the Gaussian proxy is fast, we adopt the non-parametric \(k\)-nearest-neighbour (kNN) estimator \citep{van1988generalized} for unbiased evaluation.

\noindent\textbf{Differential entropy.}
For a continuous random vector \(X\) with density \(f\),
\begin{equation}
  h(X) = -\!\int f(x)\,\log f(x)\,dx .
  \label{eq:diff_entropy}
\end{equation}

\noindent\textbf{$k$-NN estimator.}
Given \(n\) i.i.d.\ samples, let \(r_{d,k}(x_i)\) be the distance from \(x_i\) to its \(k\)th neighbour in \(\mathbb{R}^d\), and \(V_d\) the unit-ball volume.  
Then
\begin{equation}
  \hat{H}(X,k)
  \;=\;
  -\psi(k)+\psi(n)+\log V_d
  +\frac{d}{n}\sum_{i=1}^n\log r_{d,k}(x_i),
  \label{eq:knn_entropy}
\end{equation}
where \(\psi(\cdot)\) is the digamma function and \(\psi(1)=-\gamma\) with the Euler–Mascheroni constant \(\gamma\approx0.5772\).  
We default to \(k=1\) (the Kozachenko–Leonenko estimator) unless stated.

\noindent\textbf{Inter-layer entropy drop.}
We set \(H_n\equiv\hat{H}(X_n,k)\) and \(\Delta H_n = H_{n+1}-H_n\).  
These drops feed directly into the variance penalty \(L_b\) defined below.

\subsection{Loss Function for Stable Compression}
\label{sec:eloss}

Our backbones are organised into \(M\) repeated blocks, each containing \(N\) near-identical sub-encoders.  
Within block \(b\in\{1,\dots,M\}\) let
\(
\Delta H_{b,n}=H_{b,n+1}-H_{b,n}
\)
denote the entropy drop of sub-encoder \(n\).

\vspace{0.4em}
\noindent\textbf{Variance penalty \(L_b\).}  
Smooth compression implies low variance among the drops inside a block:
\begin{equation}
  L_b
  \;=\;
  \frac{1}{N}\sum_{n=1}^{N}\bigl(\Delta H_{b,n}-\overline{\Delta H}_{b}\bigr)^{2},
  \label{eq:l_block}
\end{equation}
where \(\overline{\Delta H}_{b}\) is the mean drop of block \(b\).  
At optimum, \(L_b\to0\), indicating identical per-layer compression.

\vspace{0.4em}
\noindent\textbf{Block-level divergence \(D_b\).}  
We scale the penalty with a single hyper-parameter \(\lambda\):
\begin{equation}
  D_b = \lambda\,L_b .
  \label{eq:divergence}
\end{equation}

\vspace{0.4em}
\noindent\textbf{Global regularizer \(E_{\text{loss}}\).}  
The final objective sums divergences across all blocks:
\begin{equation}
  E_{\text{loss}}
  \;=\;
  \sum_{b=1}^{M} D_b .
  \label{eq:eloss_total}
\end{equation}
During training, \(E_{\text{loss}}\) acts as an information-flow regularizer: it sharpens layer-wise interpretability within each compression block while leaving task-specific heads untouched.

\noindent\textbf{Scope and limitations.}
Because the formulation assumes repeated sub-encoders of comparable capacity, its influence is localised to such structures (e.g.\ CNN stages, transformer layers).  
Our experiments later quantify this boundary and show that \(E_{\text{loss}}\) complements—rather than replaces—the primary task loss \(L\).

\section{Experiments}

\noindent\textbf{Datasets and Experimental Setup}
We evaluate our method on \textbf{KITTI} and \textbf{nuScenes}.  
The KITTI benchmark—jointly released by Karlsruhe Institute of Technology and Toyota Technological Institute at Chicago—remains one of the most widely used datasets for intelligent-driving perception \citep{geiger_2012_are,geiger2013vision}.  
It provides multi-modal data (RGB/grayscale stereo pairs, LiDAR point clouds, GPS/IMU) collected in urban, rural, and highway scenes.  
Following standard practice, we use the official split with \textbf{7\,481} training samples and \textbf{7\,518} test samples.

nuScenes extends KITTI in both scale and sensing diversity.  
Each of its \textbf{1\,000} twenty-second scenes is captured with six surround cameras, one 32-beam LiDAR, five radars, GPS, and IMU.  
The dataset contains \textbf{1.4 M} images, \textbf{390 K} LiDAR sweeps, and \textbf{1.4 M} human-annotated 3D boxes spanning 23 object categories—roughly seven times the annotation volume of KITTI.  
Key frames are sampled at 2 Hz (40 per scene) and fully labelled; intermediate sweeps are provided without annotation.

\noindent\textbf{Implementation details.}
All experiments are run on a cluster with eight NVIDIA RTX 4090 GPUs with over 100 hours of computation time.  
Models are implemented in \textsc{PyTorch} and trained with the \textsc{MMDetection3D} toolbox \citep{mmdet3d2020}, whose data loaders, evaluation scripts, and baseline backbones we reuse.  
Unless otherwise noted, we keep the toolbox’s default training schedules and add our information-flow regularizer with a single weight \(\lambda = 1.0\), summed with the standard detection loss.

\noindent\textbf{Evaluation metrics.}
For 3D object detection on KITTI we report Average Precision under the $R_{40}$ protocol (AP$_{R40}$) for the Car, Pedestrian, and Cyclist classes on the Easy, Moderate, and Hard splits.  
On nuScenes we follow the official evaluation and report:  
(i) mean Average Precision (mAP) across the 10 object categories,  
(ii) the nuScenes Detection Score (NDS), a composite metric that combines mAP with five True Positive metrics (translation, scale, orientation, velocity, and attribute error), and  
(iii) per-class AP at a 1 m center-distance threshold (AP$_{\text{dist}1.0}$).  
All results are computed using the MMDetection3D toolkit’s built-in evaluators.

\subsection{Sensitivity of \texorpdfstring{$E_{\text{loss}}$}{Eloss} to Abnormal Inputs}
\label{sec:sensitivity}

\begin{table*}[ht]
  \centering
  \resizebox{\textwidth}{!}{
  \begin{tabular}{l|c|ccc|ccc|ccc}
    \toprule
    \multirow{2}{*}{Model \& Dataset} & \multirow{2}{*}{Value} &
    \multicolumn{3}{c|}{Confidence (no $E_{\text{loss}}$)} &
    \multicolumn{3}{c|}{$E_{\text{loss}}$ (metric only)} &
    \multicolumn{3}{c}{$E_{\text{loss}}$ (metric + loss)}\\
     &   & clean & noise1 & noise2 & clean & noise1 & noise2 & clean & noise1 & noise2 \\
    \midrule
    \multirow{2}{*}{VoxelNet KITTI} & Mean
          & 0.495 & 0.248 & 0.248 & 0.015 & 0.008 & 0.009 & 1.58E$-$3 & 9.09E$-$3 & 8.70E$-$3\\
    & \%$\Delta$
          & 0.0 & -49.9 & -49.9 & 0.0 & -48.5 & -39.1 & 0.0 & \textcolor{ForestGreen}{\textbf{+473.5}} & \textcolor{ForestGreen}{\textbf{+449.0}}\\
    \midrule
    \multirow{2}{*}{PointPillars KITTI} & Mean
          & 0.487 & 0.344 & 0.344 & 0.012 & 2.09 & 0.008 & 1.09E$-$1 & 3.84 & -- \\
    & \%$\Delta$
          & 0.0 & -29.3 & -29.3 & 0.0 & \textcolor{ForestGreen}{\textbf{+17476}} & -36.1 & 0.0 & \textcolor{ForestGreen}{\textbf{+3416}} & -- \\
    \midrule
    \multirow{2}{*}{PointPillars nuScenes} & Mean
          & 0.168 & 0.128 & 0.128 & 0.034 & 1.92 & 0.016 & 2.56E$-$4 & 1.48E$-$1 & 4.27E$-$3\\
    & \%$\Delta$
          & 0.0 & -23.7 & -23.7 & 0.0 & \textcolor{ForestGreen}{\textbf{+5495}} & -54.5 & 0.0 & \textcolor{ForestGreen}{\textbf{+57746}} & \textcolor{ForestGreen}{\textbf{+1572}}\\
    \bottomrule
  \end{tabular}}
    \caption{
    Sensitivity to abnormal inputs. Two corruption types were tested: \textbf{noise1} applies additive Gaussian noise to each point coordinate, while \textbf{noise2} applies salt–pepper noise.  
    \textbf{Confidence} refers to the mean softmax score of the top predicted class.  
    \textbf{\%$\Delta$} shows the relative change in confidence under each condition, computed as the percentage difference with respect to the corresponding \texttt{clean} value in each column.  
    For example, \%$\Delta = |\frac{\text{confidence}_{\text{noise}} - \text{confidence}_{\text{clean}}}{\text{confidence}_{\text{clean}}}| \times 100\%$.  
    }
  \label{abnormal_detection_performance}
\end{table*}

\noindent\textbf{Noise generation.}  
We simulate two distinct corruptions on LiDAR frames:  
\textbf{noise1} adds Gaussian noise sampled from a standard normal distribution to each point coordinate;  
\textbf{noise2} applies salt–pepper noise by randomly setting points to either the frame’s maximum or minimum coordinate value.

\noindent\textbf{Protocol}  
We train PointPillars and VoxelNet with and without the variance-based regularizer, then evaluate on clean KITTI and nuScenes test sets and on the two noisy variants above.  
For each configuration we report:  
(i) \textbf{Confidence}—mean classification score from the baseline network;  
(ii) \textbf{$E_{\text{loss}}$ (metric only)}—post-hoc entropy variance on the baseline;  
(iii) \textbf{$E_{\text{loss}}$ (metric + loss)}—entropy variance of the model trained with the regularizer.

\noindent\textbf{Findings}
\begin{itemize}
  \item \textbf{Confidence is weakly sensitive.}  Noise lowers confidence modestly (up to –29\%), making it hard to distinguish corruption types or levels.
  \item \textbf{$E_{\text{loss}}$ is highly sensitive.}  Even as a post-hoc metric it increases by orders of magnitude under noise1, and changes distinctly under noise2.
  \item \textbf{Training with $E_{\text{loss}}$ amplifies separation.}  With the regularizer, noise-induced spikes grow even larger (e.g.\ +57\,700\% on nuScenes for noise1), while confidence remains flat.
\end{itemize}

\noindent\textbf{Conclusion}  
$E_{\text{loss}}$ provides a robust, architecture-agnostic signal for distinguishing clean, Gaussian-noised, and salt–pepper–noised inputs, making it an effective indicator of input quality.

\subsection{Impact of \texorpdfstring{$E_{\text{loss}}$}{Eloss} on Multiple Architectures and Metrics}
\label{sec:model_comparison}

To study how the regularizer interacts with model capacity and with LiDAR–RGB fusion, we freeze the \textbf{same} voxel encoder (identical to VoxelNet) and compare three architectures of increasing complexity:

\begin{enumerate}[label=(\alph*),leftmargin=2em]
  \item \textbf{SECOND}: LiDAR-only baseline \citep{yan2018second};
  \item \textbf{SECOND+ResNet}: early fusion of LiDAR voxels with RGB features from a ResNet-50 backbone \citep{anand2024enhanced};
  \item \textbf{SECOND+FusionStack}: the previous model augmented with a correlation module \citep{zheng2022multi}, a GNN head \citep{lei2024gnn}, and an FPN neck \citep{lin2017feature}.
\end{enumerate}

Each model is fine-tuned for 40 epochs \textbf{with} or \textbf{without} $E_{\text{loss}}$ (inserted into all SECOND blocks) and then evaluated on KITTI.  
Table~\ref{tab:model_comparison} lists Car / Cyclist / Pedestrian $AP_{R40}$ and the percentage change ($\Delta$).

\begin{table*}[h]
  \centering
  \resizebox{0.99\textwidth}{!}{
  \begin{tabular}{l|l|ccc|ccc|ccc}
    \toprule
    \multirow{2}{*}{Architecture} & \multirow{2}{*}{Value} &
      \multicolumn{3}{c|}{Car} &
      \multicolumn{3}{c|}{Cyclist} &
      \multicolumn{3}{c}{Pedestrian} \\
    & & Easy & Mod. & Hard & Easy & Mod. & Hard & Easy & Mod. & Hard \\
    \midrule
    \multirow{3}{*}{SECOND} 
      & base   & 82.35 & 73.35 & 68.59 & 70.89 & 56.72 & 50.68 & 50.75 & 40.76 & 36.96 \\
      & $E_{\text{loss}}$ & 82.68 & 73.67 & 67.21 & 71.99 & 58.00 & 50.94 & 50.49 & 41.16 & 37.43 \\
      & \%$\Delta$ & \textcolor{ForestGreen}{\textbf{+0.33}} & \textcolor{ForestGreen}{\textbf{+0.32}} & -1.38 & \textcolor{ForestGreen}{\textbf{+1.10}} & \textcolor{ForestGreen}{\textbf{+1.28}} & \textcolor{ForestGreen}{\textbf{+0.26}} & -0.26 & \textcolor{ForestGreen}{\textbf{+0.40}} & \textcolor{ForestGreen}{\textbf{+0.47}} \\
    \midrule
    \multirow{3}{*}{SECOND\,{+}\,ResNet} 
      & base   & 80.29 & 67.37 & 60.94 & 75.70 & 52.37 & 46.10 & 39.64 & 31.13 & 28.95 \\
      & $E_{\text{loss}}$ & 77.62 & 64.92 & 60.36 & 71.47 & 55.79 & 49.64 & 44.85 & 35.60 & 32.66 \\
      & \%$\Delta$ & -2.67 & -2.45 & -0.58 & -4.23 & \textcolor{ForestGreen}{\textbf{+3.42}} & \textcolor{ForestGreen}{\textbf{+3.54}} & \textcolor{ForestGreen}{\textbf{+5.21}} & \textcolor{ForestGreen}{\textbf{+4.47}} & \textcolor{ForestGreen}{\textbf{+3.71}} \\
    \midrule
    \multirow{3}{*}{SECOND\,{+}\,FusionStack} 
      & base   & 73.47 & 62.47 & 57.99 & 63.08 & 49.55 & 44.33 & 42.46 & 35.11 & 32.16 \\
      & $E_{\text{loss}}$ & 67.33 & 58.70 & 54.13 & 57.16 & 46.36 & 41.54 & 45.02 & 36.39 & 33.29 \\
      & \%$\Delta$ & -6.14 & -3.77 & -3.86 & -5.92 & -3.19 & -2.79 & \textcolor{ForestGreen}{\textbf{+2.56}} & \textcolor{ForestGreen}{\textbf{+1.28}} & \textcolor{ForestGreen}{\textbf{+1.13}} \\
    \bottomrule
  \end{tabular}}
  \caption{KITTI $AP_{R40}$ (\%) after 40-epoch fine-tune with or without $E_{\text{loss}}$. Positive $\Delta$ means an improvement.}
  \label{tab:model_comparison}
\end{table*}

\noindent\textbf{Analysis}
\begin{itemize}[leftmargin=1.3em]
  \item \textbf{SECOND.}  Adding $E_{\text{loss}}$ gives small gains for Car (Easy/Mod) and Cyclist and keeps Pedestrian almost unchanged.
  \item \textbf{SECOND+ResNet.}  Early LiDAR-RGB fusion boosts minor classes (Cyclist and Pedestrian, up to +5\%) but reduces Car accuracy, suggesting that $E_{\text{loss}}$ suppresses noisy image cues mostly useful for large, frequent objects.
  \item \textbf{SECOND+FusionStack.}  Deeper fusion magnifies the trade-off: Pedestrian improves while Car and Cyclist drop.  Heavy post-fusion processing may distort entropy-stabilized features unless outer-layer learning rates are reduced; we leave this for future work.
\end{itemize}

In summary, $E_{\text{loss}}$ often helps smaller or rarer categories once the network has enough capacity or multi-modal context, but a strong variance constraint can hurt dominant classes when added after extensive fusion.
\subsection{Emergent Monotonic Compression in Feature Space}
\label{app:eloss_distribution}

\begin{figure}[ht]
  \centering
  \includegraphics[width=\linewidth]{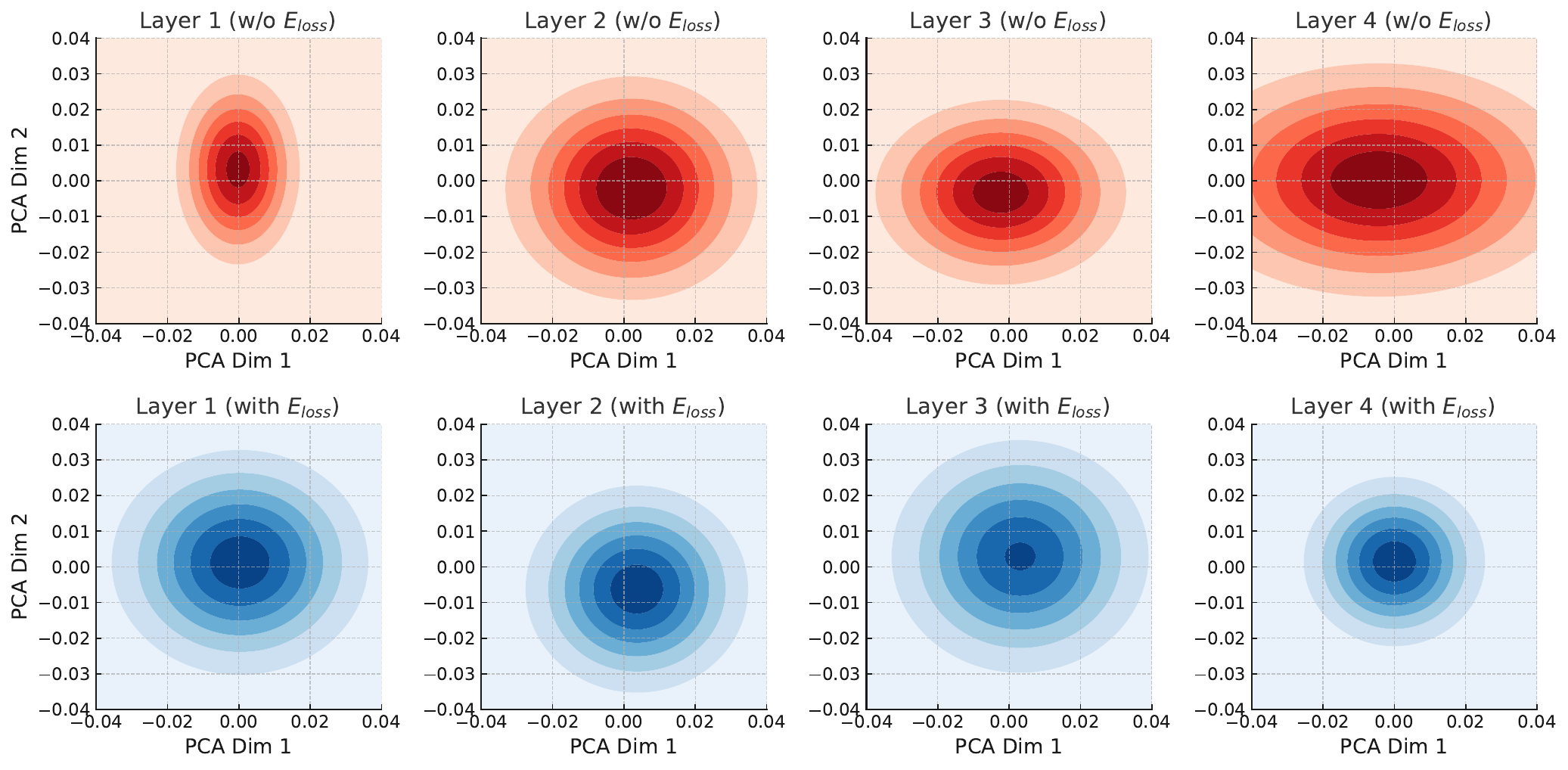}
  \caption{Comparison of layer-wise feature distributions in PCA space with and without $E_{\text{loss}}$.  
  Each subplot shows the product of two 1D Gaussians fitted separately to the first two principal components of normalized feature activations from the SECOND backbone.  
  The contour levels visualize the approximated 2D density.  
  \textbf{Top row:} Layers 1–4 without $E_{\text{loss}}$ exhibit irregular or outward-drifting energy spread.  
  \textbf{Bottom row:} The same layers with $E_{\text{loss}}$ display progressively contracting, nearly spherical contours, indicating smoother and more stable compression.  
  All plots use the same axis limits ($\pm 0.04$) and square aspect ratio for comparability.}
  \label{fig:eloss_distribution}
\end{figure}

To examine how $E_{\text{loss}}$ reshapes internal representations, we extract the output from each layer of a single SECOND block and project the normalized feature tensors onto their first two principal components, visualized via density contours in Figure~\ref{fig:eloss_distribution}.

\noindent\textbf{Observation}  
Without $E_{\text{loss}}$, the feature distributions remain irregular and expand outward layer-by-layer, suggesting unstable information flow.  
In contrast, models trained with $E_{\text{loss}}$ yield feature activations that progressively contract toward the origin and exhibit near-spherical density shapes—especially in deeper layers.

\noindent\textbf{Interpretation}  
These spherical contours reflect the effect of the $k$-nearest-neighbour entropy estimator (Equation~\ref{eq:knn_entropy}), which is most stable under isotropic, compact distributions.  
The visual contraction across layers confirms that enforcing low variance in entropy drops (Principle D1) is sufficient to induce monotonic compression (Principle D2), without requiring an explicit monotonicity constraint.

\noindent\textbf{Implication}
Figure \ref{fig:eloss_distribution} visually confirms that enforcing low‐variance entropy drops produces not only a gradual, layer‐by‐layer reduction in information spread but also an even, direction-neutral contraction of feature activations. This uniform shaping improves the reliability of k-NN entropy estimates and yields more robust, stable representations. 

\section{Conclusion}
We introduced \textbf{E\textsubscript{loss}}, a novel information-theoretic regularizer designed to enforce stable and interpretable information flow within deep neural perception backbones. Our approach reframes deep perception networks as hierarchical communication systems, guiding training toward smooth, consistent compression of raw sensory inputs into task-relevant latent representations. By explicitly penalizing the variance of entropy drops between layers, \textbf{E\textsubscript{loss}} leverages foundational principles of information theory, resulting in monotonic entropy decay and enhanced interpretability without the need for explicit monotonic constraints.

Experiments on the large-scale autonomous driving benchmarks KITTI and nuScenes demonstrate that \textbf{E\textsubscript{loss}} consistently achieves comparable or improved detection accuracy relative to standard baselines. More importantly, our method significantly increases model sensitivity to anomalous inputs—boosting detection signals for distribution shifts by up to two orders of magnitude. These outcomes reflect not merely incremental accuracy gains but a fundamental shift toward robust, theory-grounded perception systems.

Our work emphasizes that stable, interpretable information processing is both achievable and beneficial for safety-critical applications like autonomous driving. Future research directions include adapting \textbf{E\textsubscript{loss}} to networks with heterogeneous layer capacities, further reducing computational overhead, and extending our framework to other challenging perception tasks and real-world operational environments.


{
\small
\bibliographystyle{unsrtnat}
\bibliography{main}
}


\appendix


\section{Comparison with Eloss on Training Process}
\label{sec:eloss_training}
    \begin{figure*}[ht!]
      \centering
      \subfloat[Car $AP_{dist1.0}$]{
        \includegraphics[width=0.333\textwidth]{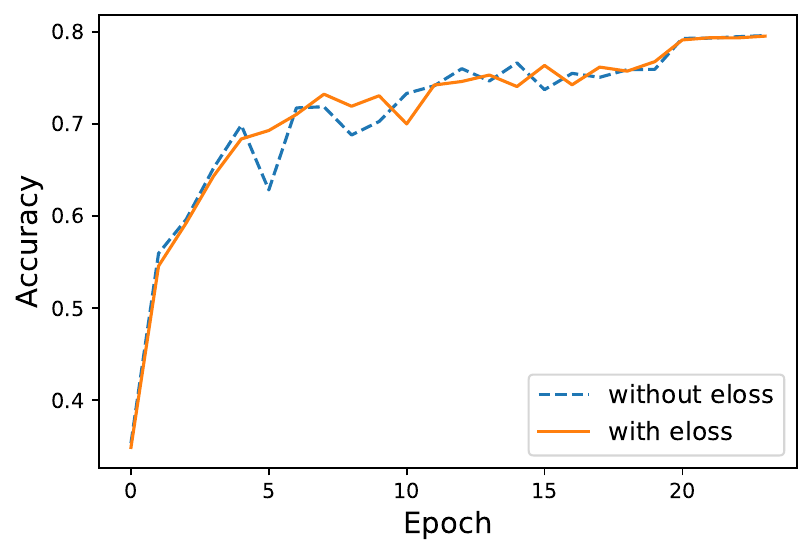}
      }
      \subfloat[mAP]{
        \includegraphics[width=0.333\textwidth]{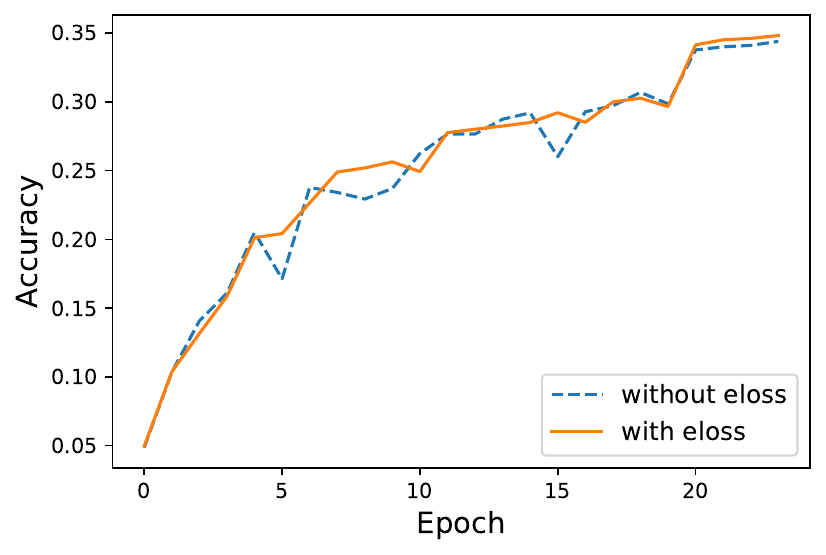}
      } 
      \subfloat[NDS]{
        \includegraphics[width=0.333\textwidth]{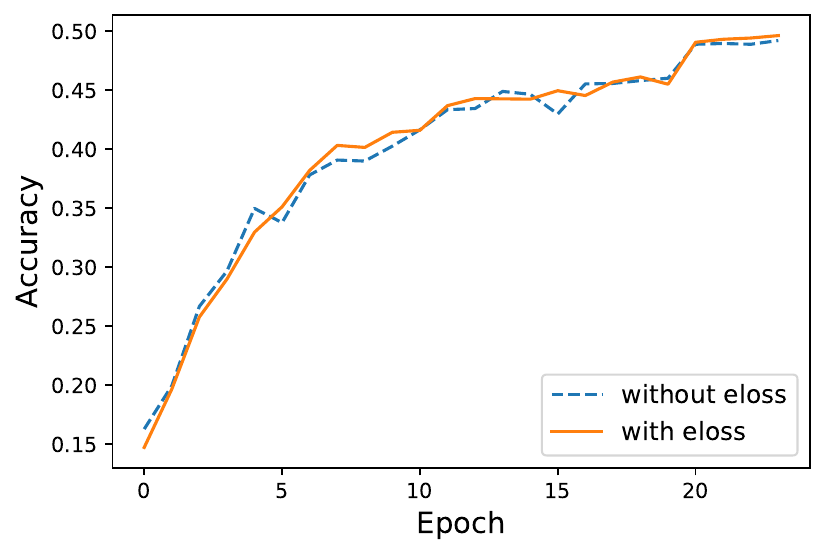}
      } 
      \caption{Convergences Curves of the model accuracy on NuSenes validation set for PointPillars  with or without Eloss. (a) the Average Precision of Car detection with Distance Threshold 1.0 meters; (b) mean Average Precision computed across 10 class of objects; (c) nuScenes detection score.}
      \label{training_process_comparison_figure}
    \end{figure*}

    To measure the impact of eloss on the model training process, we first conduct control experiments on the same model with and without eloss on the KITTI dataset and Nusenes dataset without noise. We plot the part of our experiment results in Figure~\ref{training_process_comparison_figure} to more intuitively show the impact of eloss on the volatility of the training process.
    
    To quantify the impact process, we use the mean absolute value slope (Mean Absolute Value Slope, MAVP) to measure the impact of eloss on the volatility of the model training curve and use the maximum precision index to measure the impact of eloss on the model training accuracy. The MAVP formula is as follows, where N is the number of sliding panes, and (k, k+1) refers to two adjacent time windows.

    \begin{equation}
        \mbox{MAVP}=\frac{1}{N}\sum_{k=1}^N(|x_{k+1}|-|x_k|))
    \end{equation}

    We applied the pointpillars and voxelnet models to the KITTI dataset to conduct the above control experiments. The experimental results are shown in the Table~\ref{kitti_eval_table}.

    \begin{table*}[ht!]
    \centering
    \begin{minipage}{0.48\textwidth}
        \centering
        \resizebox{\textwidth}{!}{
            \begin{tabular}{l|l|cc}
                \hline
                Model & Method & Max(\%) & MAVP(\%) \\ \hline
                \multirow{3}{*}{PointPillars} & Without Eloss & 90.694 & 11.946 \\ 
                ~ & With Eloss & 88.916 & 11.932 \\ 
                ~ & Delta & -1.778 & \textbf{-0.014} \\ \hline
                \multirow{3}{*}{VoxelNet} & Without Eloss & 94.873 & 10.959 \\  
                ~ & With Eloss & 94.586 & 10.937 \\
                ~ & Delta & -0.287 & \textbf{-0.022} \\ \hline
            \end{tabular}
        }
        \caption{Car $AP_{R40}$ Max and MAVP on KITTI validation set during training.}
        \label{kitti_eval_table}
    \end{minipage}
    \hfill
    \begin{minipage}{0.48\textwidth}
        \centering
        \resizebox{\textwidth}{!}{
            \begin{tabular}{l|l|cc}
                \hline
                Metric & Method & Max(\%) & MAVP(\%) \\\hline
                \multirow{3}{*}{Car $AP_{dist 1.0}$} & Without Eloss & 79.580 & 2.945 \\ 
                ~ & With Eloss & 79.520 & 2.811 \\ 
                ~ & Delta & -0.060 & \textbf{-0.135} \\ \hline
                \multirow{3}{*}{mAP} & Without Eloss & 34.393 & 6.036 \\ 
                ~ & With Eloss & 34.815 & 4.883 \\
                ~ & Delta & \textbf{0.422} & \textbf{-1.153} \\ \hline
                \multirow{3}{*}{NDS} & Without Eloss & 49.217 & 4.902 \\ 
                ~ & With Eloss & 49.637 & 3.902 \\ 
                ~ & Delta & \textbf{0.420} & \textbf{-1.000} \\ \hline
            \end{tabular}
        }
        \caption{Max and MAVP of PointPillars on nuScenes validation set during training.}
        \label{nus_eval_table}
    \end{minipage}
\end{table*}

    The experimental results in the pivot table show that the maximum training accuracy decreases after adding eloss to both models. In terms of MAVP, the MAVP decreased after adding eloss, which means that the addition of eloss makes the above training process smoother.

    On the Nuscence dataset, we perform the above control experiments on the PointPillars model with three different metrics: Car $AP_dist1.0$, mAP, and NDS. The experimental results are shown in the Table~\ref{nus_eval_table}.

    The table shows that the maximum accuracy of the Car category is lost during the training process after adding eloss. Still, the decline in MAVP shows that the addition of eloss moderates the volatility of the above training process. Similar observations for mAP, the average precision of multiple categories, and NDS, the Nusenes detection score, indicate that adding eloss to the model makes the training process smoother.

    The above is the experiment on the influence of eloss on model training without noise interference. In order to further understand the effect of eloss in the experiment, we will add Eloss to different parts of the network or conduct control experiments with anomalous data.

\section{PointPillar Sensitivity to $E_{\text{loss}}$ Coverage}
\label{sec:eloss_coverage}

We resume training PointPillars models that have completed 80 epochs and inject $E_{\text{loss}}$ into an increasing number of SECOND blocks.\footnote{All other hyper-parameters remain fixed; only the number of blocks governed by $E_{\text{loss}}$ is varied.}  
Because the regularizer is fully plug-and-play, it can be attached to any repetitive sub-structure.  Here we vary its coverage from 0 to 3 blocks.

\begin{figure}[ht!]
  \centering
  \includegraphics[width=0.732\linewidth]{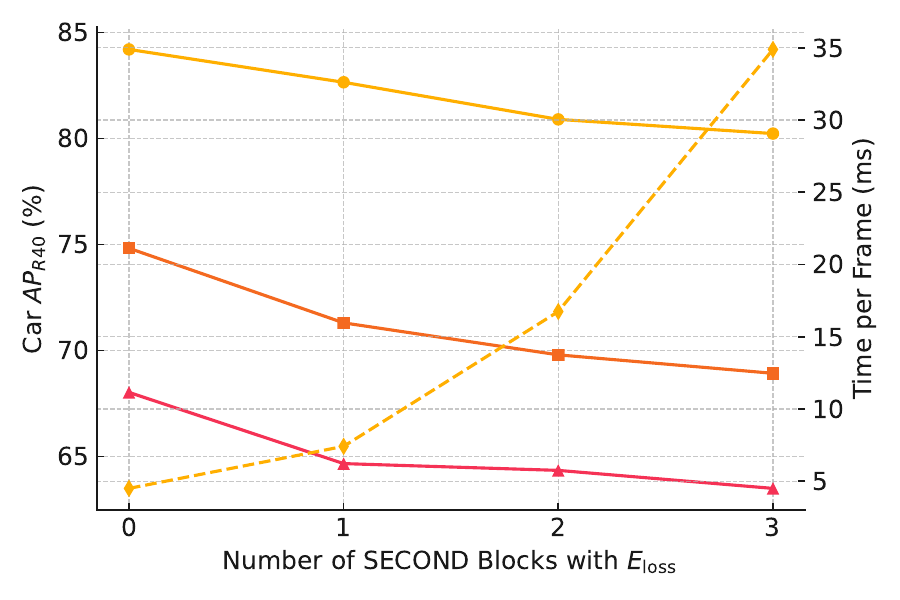}
  \caption{PointPillars on KITTI with $E_{\mathrm{loss}}$ applied to 0–3 SECOND blocks.  
  \textbf{Circles} = Easy, \textbf{Squares} = Moderate, \textbf{Triangles} = Hard (left axis).  
  \textbf{Diamonds dashline} = per-frame inference time (right axis).}
  \label{fig:eloss_coverage}
\end{figure}

\noindent\textbf{Observation.}
As the regularizer constrains more blocks, all three difficulty levels of Car $AP_{R40}$ steadily decline, while runtime rises from $4.5$ ms (0 blocks) to $34.9$ ms (3 blocks).  
This corroborates the hypothesis that tighter entropy-stability constraints narrow the optimization landscape: the detector sacrifices peak accuracy in exchange for stronger robustness priors and incurs additional computation to evaluate the entropy terms.

\noindent\textbf{Remark on VoxelNet.}
Repeating the coverage study on VoxelNet produces an {interesting} non-monotonic pattern: accuracy first dips when one block is regularized, but rebounds—and even surpasses the baseline—when two blocks are constrained.
The complete results and discussion are presented in Appendix~\ref{sec:voxelnet_eloss}.

\section{VoxelNet Sensitivity to \texorpdfstring{$E_{\text{loss}}$}{Eloss} Coverage}
\label{sec:voxelnet_eloss}

To verify whether the coverage pattern observed for PointPillars generalises, we repeat the block-injection study on VoxelNet.

\begin{table}[h]
  \centering
  \resizebox{0.8\columnwidth}{!}{
  \begin{tabular}{c c c |ccc| c}
    \toprule
    Model & Epoch & \#Blocks w/ $E_{\text{loss}}$ &
    \multicolumn{3}{c|}{Car $AP_{R40}$ (\%)} & Time (ms)\\
          &       &           & Easy & Mod. & Hard & \\
    \midrule
    \multirow{3}{*}{VoxelNet}
      & 80 & 0 & 82.35 & 73.35 & 68.59 & 6.94\\
      & 85 & 1 & 81.34 & 70.46 & 65.29 & 14.72\\
      & 85 & 2 & 85.34 & 74.44 & 67.51 & 33.75\\
    \bottomrule
  \end{tabular}}
  \caption{VoxelNet on KITTI with $E_{\text{loss}}$ applied to 0–2 SECOND blocks.}
  \label{tab:eloss_amount_voxelnet}
\end{table}

\noindent\textbf{Discussion.}
Contrary to PointPillars (Figure~\ref{fig:eloss_coverage}), VoxelNet shows a \textit{non-monotonic} trend: injecting $E_{\text{loss}}$ into a single block reduces all three AP scores, yet constraining {two} blocks surpasses the baseline on Easy and Moderate.  
We posit two factors:

\begin{itemize}
\item \textbf{Receptive-field size.} VoxelNet’s 3D sparse convolutions already cover large spatial extents; a stronger stability prior may curb over-fitting to high-frequency noise, yielding a net gain when applied to multiple blocks.
\item \textbf{Normalisation effects.} Preliminary ablations (omitted for space) indicate that layer-norm placement modulates the strength of the entropy signal; one-block regularisation may under- or over-weight this interaction.
\end{itemize}

A comprehensive sweep—including per-block ablations and alternative normalisation layouts—is left for future work.

\end{document}